%% file: HARP.tex
\documentclass{article}
\usepackage[T1]{fontenc}
\usepackage{newtxtext}  
\usepackage{spconf,amsmath,amssymb,graphicx}
\usepackage{booktabs,multirow}
\usepackage[table]{xcolor}
\usepackage{indentfirst}
\usepackage[final]{microtype}

\usepackage[pass]{geometry}
\makeatletter
\let\oldthebibliography\thebibliography
\let\endoldthebibliography\endthebibliography
\renewenvironment{thebibliography}[1]{%
  \oldthebibliography{#1}%
  \fontsize{9pt}{9.5pt}\selectfont
  \setlength{\itemsep}{0.1ex}
  \setlength{\parsep}{1ex}
  \setlength{\parskip}{0pt}%
  \setlength{\topsep}{0pt}%
  \setlength{\partopsep}{0pt}%
}{%
  \endoldthebibliography
}
\makeatother

\usepackage{etoolbox}

\usepackage[hidelinks]{hyperref}

\title{\MakeUppercase{MSP-ReID: Hairstyle-Robust Cloth-Changing Person Re-Identification}}

\name{Xiangyang He\textsuperscript{\textup{1}} \qquad Lin Wan\textsuperscript{\textup{1,*}} \thanks{*Corresponding author (wanlin@cug.edu.cn).}}

\address{\textsuperscript{1} School of Computer Science, China University of Geosciences, Wuhan, China}

\begin{document}
\ninept  
\topmargin=0mm

\maketitle

\begin{abstract}

Cloth-Changing Person Re-Identification (CC-ReID) aims to match the same individual across cameras under varying clothing conditions. Existing approaches often remove apparel and focus on the head region to reduce clothing bias. However, treating the head holistically without distinguishing between face and hair leads to over-reliance on volatile hairstyle cues, causing performance degradation under hairstyle changes.
To address this issue, we propose the Mitigating Hairstyle Distraction and Structural Preservation (MSP) framework. Specifically, MSP introduces Hairstyle-Oriented Augmentation (HSOA), which generates intra-identity hairstyle diversity to reduce hairstyle dependence and enhance attention to stable facial and body cues. To prevent the loss of structural information, we design Cloth-Preserved Random Erasing (CPRE), which performs ratio-controlled erasing within clothing regions to suppress texture bias while retaining body shape and context. Furthermore, we employ Region-based Parsing Attention (RPA) to incorporate parsing-guided priors that highlight face and limb regions while suppressing hair features. Extensive experiments on multiple CC-ReID benchmarks demonstrate that MSP achieves state-of-the-art performance, providing a robust and practical solution for long-term person re-identification.

\end{abstract}


\vspace{-1ex}
\section{Introduction}
\label{sec:intro}
Person re-identification (Re-ID) aims to recognize the same individual across different cameras and time intervals~\cite{qian2017multi,wang2025distribution}. With the increasing demand for large-scale intelligent surveillance, short-term Re-ID research has become relatively mature, where a person’s clothing typically remains unchanged~\cite{wang2024large}. However, models trained under such settings tend to overfit clothing appearance, causing significant performance degradation when individuals change outfits or different people wear visually similar clothes. This limitation has motivated research on \emph{cloth-changing} person re-identification (CC-ReID), which aims to ensure robust identity matching under clothing variations and better satisfies the requirements of long-term, real-world applications~\cite{qian2020long,wang2022co,wang2024image}.

\begin{figure}[t]
  \centering
  \includegraphics[width=0.48\textwidth, height=0.26\textheight]{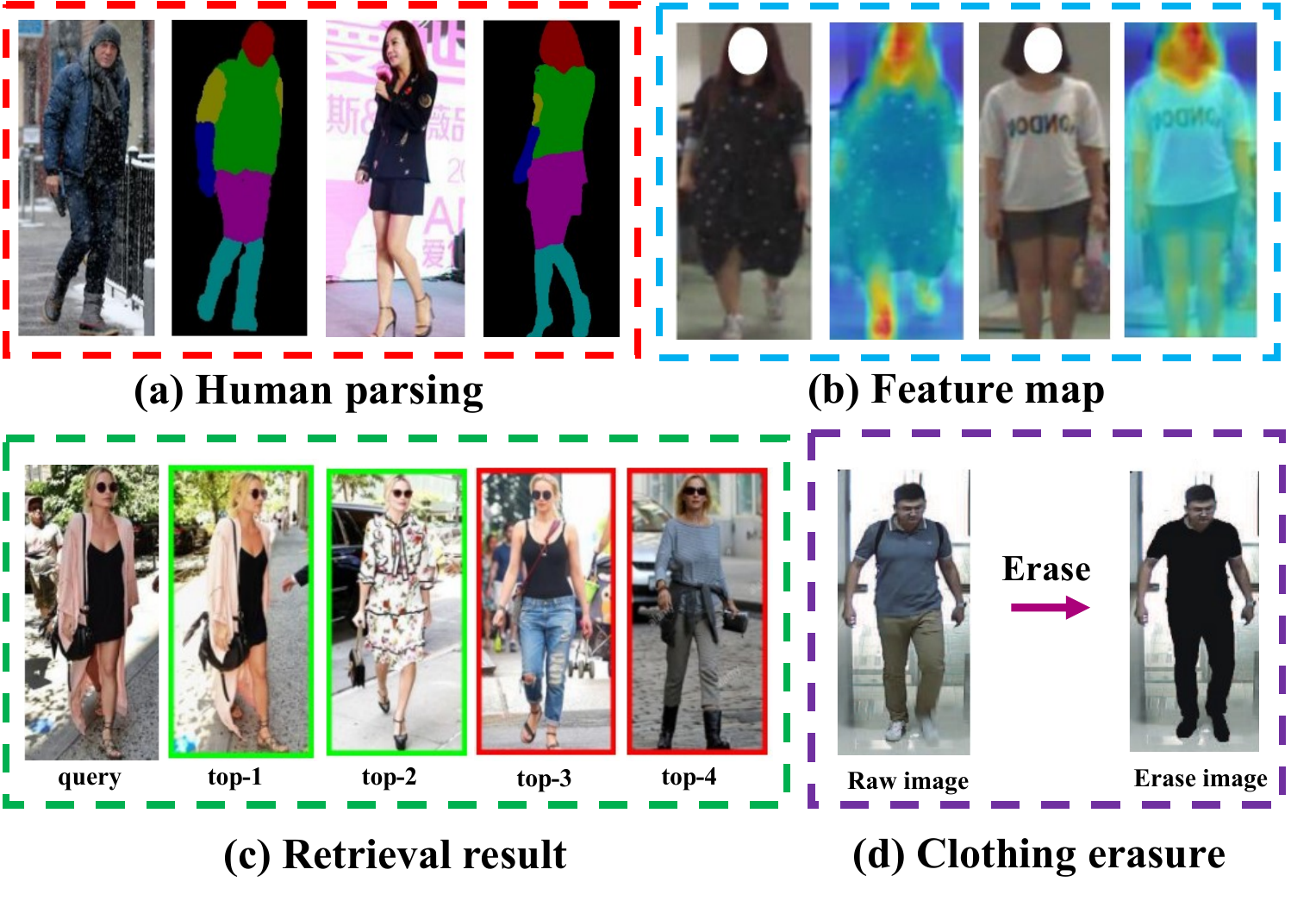}
  \vspace{-5ex}
  \caption{\textbf{The Hairstyle Shortcut problem in CC-ReID.}
  (a) Standard parsing merges face and hair together as "head";
  (b) attention consequently focuses excessively on this region;
  (c) resulting models are robust to clothing changes but brittle to hairstyle variations;
  (d) conventional clothing erasure further removes structural cues.}
  \label{fig:problem}
  \vspace{-4ex}
\end{figure}

Beyond clothing changes, factors such as hairstyle variations and aging also significantly affect appearance, yet both are visually salient but identity-irrelevant features. Existing CC-ReID methods typically reduce clothing dependence through semantic mining, feature attention, and regional erasing~\cite{qian2020long,hong2021fine,wang2024exploring,thanh2024enhancing,xiong2024cloth}, often without relying on auxiliary modalities~\cite{gu2022clothes,yang2023good,han2023clothing}. However, a critical challenge remains largely overlooked: the impact of hairstyle variations on recognition has not been adequately addressed. Standard parsing techniques generally mark the entire head (including both face and hair) as identity-related \cite{thanh2024enhancing,guo2023semantic} (Fig.~\ref{fig:problem}(a)), causing models to overemphasize head regions and become highly sensitive to hairstyle cues (Fig.~\ref{fig:problem}(b)). This introduces a "hairstyle shortcut" whereby models mistakenly rely on hairstyle as the primary identity cue, degrading their generalization ability when hairstyles change (Fig.~\ref{fig:problem}(c)).Furthermore, parsing-based clothing removal methods are often too aggressive~\cite{guo2023semantic}, eliminating not only clothing pixels but also crucial structural cues—such as body silhouette, proportions, and pose—further weakening the model"s generalizability (Fig.~\ref{fig:problem}(d)).

\begin{figure*}[t]
  \centering
  \includegraphics[width=0.86\textwidth, height=0.33\textheight]{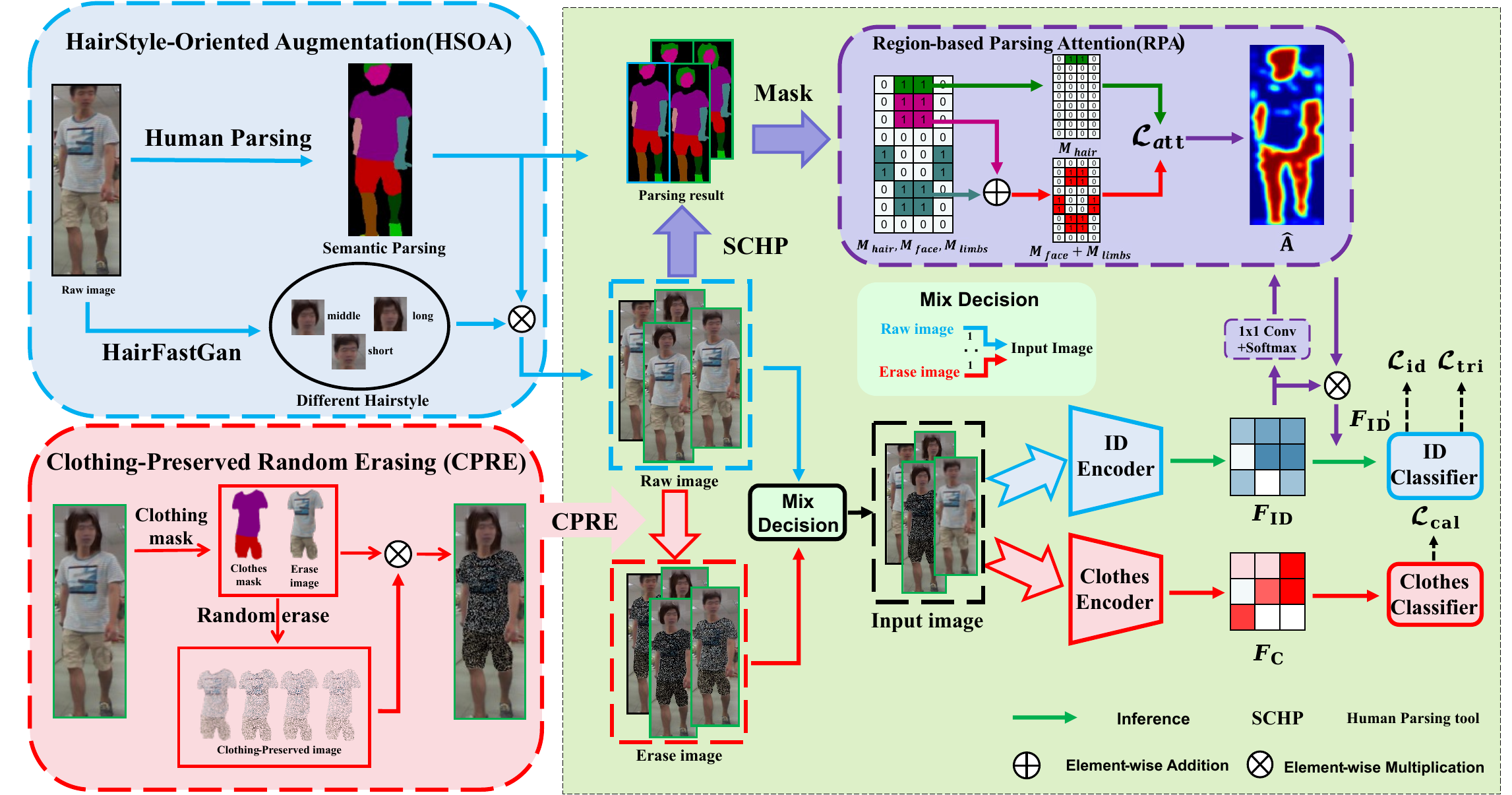} 
  \caption{Overview of MSP-ReID. \textbf{HSOA} (blue dashed, top-right) synthesizes same-ID different-hairstyle images. \textbf{CPRE} (pink dashed, bottom-right) creates raw/erased pairs with a clothing keep ratio. \textbf{RPA} (purple, center-top) uses parsing masks to boost face/limbs and suppress hair. Green denotes the ID branch, pink denotes the clothes branch for adversarial regularization. Inference is RGB-only using the ID branch.}
  \label{fig:framework}
  \vspace{-2ex}
\end{figure*}

Motivated by these challenges, we propose the Mitigating Hairstyle Distraction and Structural Preservation (MSP) framework to simultaneously address these limitations of existing CC-ReID methods: the over-reliance on hairstyle cues and the loss of structural information caused by complete clothing removal. Specifically, we introduce Hairstyle-Oriented Augmentation (HSOA) to generate "same-identity, different-hairstyle" samples and align their features in the embedding space, explicitly decoupling hairstyle from identity representation. To preserve body geometry, we design Cloth-Preserved Random Erasing (CPRE), which retains a controllable portion of clothing pixels, suppressing texture bias while maintaining body contours, posture, and proportions. Finally, we propose Region-based Parsing Attention (RPA) to leverage human parsing priors, strengthening identity-relevant regions (e.g., face, limbs) and suppressing non-identity features such as hair.

\noindent\textbf{Contributions.} The contributions are summarized as follows.

(1) We pioneer MSP-ReID, a framework that explicitly addresses hairstyle-induced bias for the first time, leading to consistent improvement in robustness and performance in CC-ReID.

(2) We present Hairstyle-Oriented Augmentation (HSOA) to decouple hairstyle cues from identity learning and propose Region-based Parsing Attention (RPA) to focus representation on stable facial and body regions while suppressing hair-induced noise.

(3) We propose Cloth-Preserved Random Erasing (CPRE), which differs from conventional random erasing by maintaining a controllable ratio of clothing pixels, preserving geometric information (body shape, pose) and suppressing texture dependency.

(4) Extensive experiments confirm MSP’s robustness and effectiveness, achieving a new state-of-the-art for CC-ReID.

\vspace{-1ex}
\section{Methodology}
\label{sec:method}

\subsection{Problem Formulation}
Let $x\!\in\!\mathbb{R}^{H\times W\times 3}$ be a pedestrian image with identity label $y$.
Besides identity, the appearance also contains time-varying \emph{nuisance} attributes such as clothing $c$ and hairstyle $h$.
We posit an underlying data-generating distribution $p(x,y,c,h)$ but only observe $(x,y)$ during training, $(c,h)$ are unannotated.
Our goal is to learn an encoder $f_\theta$ that maps $x$ to an embedding $z=f_\theta(x)$ which is (i) highly discriminative for identity and (ii) insensitive to $(c,h)$.
Formally, we seek a representation $z$ that maximizes identity information while minimizing dependence on nuisance factors:
\begin{equation}
\max_{\theta}\, \{I\!\big(f_\theta(X),Y\big)\,-\,\lambda_c\,I\!\big(f_\theta(X),C\big)\,-\,\lambda_h\,I\!\big(f_\theta(X),H\big)\},
\end{equation}
where $I(\cdot,\cdot)$ denotes mutual information, i.e., the amount of information one variable reveals about the other
($I(Z,C)=0$ implies $Z$ and $C$ are statistically independent).
Here, $Y$ denotes the identity label, $C$ the clothing state, and $H$ the hairstyle state.
Equivalently, we aim for $z$ to preserve identity-discriminative cues (large inter-identity margins, low intra-identity variance) while suppressing clothing/hairstyle cues.

\subsection{Hairstyle-Oriented Augmentation (HSOA)}
\label{ssec:hair}
To explicitly break the shortcut "hair $\approx$ identity", we perform a hairstyle augmentation with HairFastGAN~\cite{nikolaev2024hairfastgan}. For each training image $x$, we utilize the human parser SCHP~\cite{li2020self} to obtain a pixel-wise semantic map
$ P= \mathrm{SCHP}(x)\in\{1,\dots,K\}^{H\times W}$.
From $P$ we derive binary masks for face and hair and define the head mask as their union:
\begin{align}
M_{\text{face }} &= \mathbf{1}\!\left[P\in\mathcal{S}_{\text{face}}\right],\quad
M_{\text{hair}} = \mathbf{1}\!\left[P\in\mathcal{S}_{\text{hair}}\right],\\
M_{\text{head}} &= M_{\text{face}} \lor M_{\text{hair}}
= \mathbf{1}\!\left[P\in\mathcal{S}_{\text{face}}\cup\mathcal{S}_{\text{hair}}\right].
\end{align}
The cropped head region $x\odot M_{\text{head}}$ and its mask are fed to HairFastGAN to synthesize three target hairstyles—\textbf{short}, \textbf{medium}, \textbf{long}—yielding heads $\hat{h}_{\text{S}},\hat{h}_{\text{M}},\hat{h}_{\text{L}}$ that preserve facial structure while altering hair. Each synthesized head is seamlessly composited back:
\begin{equation}
\tilde{x}_\ell
= M_{\text{hair}}\odot \hat{h}_\ell
  + (1 - M_{\text{hair}})\odot x ,\quad \ell\!\in\!\{\text{S},\text{M},\text{L}\},
\end{equation}
where $\odot$ denotes element-wise multiplication. Each $\tilde{x}_\ell$ inherits the identity label $y$ and the clothing label $c$. The augmented set $\{\tilde{x}_{\text{S}},\tilde{x}_{\text{M}},\tilde{x}_{\text{L}}\}$ is sampled together with originals, creating abundant positive pairs of the \emph{same identity under different hairstyles}. Notably, we leverage triplet loss to optimize the feature space, pulling closer representations of the same identity across varying hairstyles and clothing while pushing apart those from different identities.

\subsection{Cloth-Preserved Random Erasing (CPRE)}
\label{ssec:cpre}

Although clothing contains strong visual cues, they are considered identity-unrelated features in our task.
A common practice in existing methods is to remove this information entirely to compel the model to learn identity-related features.
However, removing the entire clothing region also discards useful information about body structure and spatial context. To address this limitation, we propose Cloth-Preserved Random Erasing (CPRE), we design Cloth-Preserved Random Erasing (CPRE) to erase only within the clothing region, retaining a random proportion of clothing pixels, forcing the model to rely more on identity-related cues (face, limbs, shape). 
Let $M_{\text{cloth}}\!\in\!\{0,1\}^{H\times W}$ be the clothing mask (slightly dilated to cover boundary errors). Sample a keep ratio $r\!\in\![r_{\min},r_{\max}]$ and draw $K_r\!\in\!\{0,1\}^{H\times W}$ \emph{inside the clothing region} such that approximately a proportion $r$ is preserved (i.e., $\mathbb{E}\!\big[K_r(i,j)\,\big|\, M_{\text{cloth}}(i,j)=1\big]=r$)
. The erased image is

\begin{equation}
x^{\text{erase}}
= \Big((1-M_{\text{cloth}})+M_{\text{cloth}}\odot K_r\Big)\odot x
+ \Big(M_{\text{cloth}}\odot(1-K_r)\Big)\odot \epsilon ,
\end{equation}
where $\epsilon$ is a constant fill (zero).

\noindent\textbf{Pixel-wise image.}
Equivalently, for each pixel $(i,j)$,
\begin{equation}
x^{\text{erase}}_{i,j}=
\begin{cases}
x_{i,j}, & M_{\text{cloth}}(i,j)=0,\\
x_{i,j}, & M_{\text{cloth}}(i,j)=1\ \land\ K_r(i,j)=1,\\
\epsilon,  & M_{\text{cloth}}(i,j)=1\ \land\ K_r(i,j)=0~.
\end{cases}
\end{equation}
\noindent\textbf{image averaging.}
When CPRE is enabled, the Mix Decision module constructs input batches by the raw image $x$  and erased image $x^{\text{erase}}$ at a 1:1 ratio.

\subsection{Region-based Parsing Attention (RPA)}
\label{ssec:RPA}
While CPRE reduces reliance on clothing features, hairstyle remains a prominent, identity-unrelated distractor. To mitigate this specific problem, we propose Region-based Parsing Attention (RPA), a lightweight attention mechanism that uses human parsing priors to guide the model’s focus.It generates a spatial attention map that emphasizes identity-related regions and minimizes attention to identity-unrelated features, such as hair, helping the model learn more robust identity representations.

\noindent\textbf{Backbone and ID head.}
A backbone $B(\cdot)$ produces a feature map $F\!\in\!\mathbb{R}^{C\times H\times W}$, a shallow ID head yields $F_{\text{ID}}=\phi_{\text{id}}(F)$.

\noindent\textbf{Attention prediction and gating.}
Given $F_{\text{ID}}$, a $1\times1$ convolution predicts attention logits $S\!\in\!\mathbb{R}^{1\times H\times W}$:
\begin{equation}
S = W * F_{\text{ID}} + b,\qquad
\hat{A}_{ij} = \frac{\exp(S_{ij})}{\sum_{u,v}\exp(S_{uv})}\in(0,1),
\end{equation}
where $W\!\in\!\mathbb{R}^{1\times C\times 1\times 1}$. The gated ID features are
\begin{equation}
{F}_{\text{ID}}" = F_{\text{ID}} \odot \hat{A},
\end{equation}
with $\hat{A}$ broadcast along channels, global average pooling (GAP) of ${F}_{\text{ID}}"$ is used for downstream losses. At test time, the RPA gate is disabled and the model uses the ungated $F_{\text{ID}}$.

\noindent\textbf{Parsing-guided attention loss.}
Given the parsing masks $M_{\text{face}}, M_{\text{limbs}} \\
, M_{\text{hair}}$, with
\[
M_{\text{face}},\, M_{\text{limbs}},\, M_{\text{hair}} \in \{0,1\}^{H\times W},
\]
we define the normalized positive target
\begin{equation}
T_{\!+}=\frac{M_{\text{face}}+M_{\text{limbs}}}
{\langle \mathbf{1},\,M_{\text{face}}+M_{\text{limbs}}\rangle+\varepsilon}.
\end{equation}
We supervise $\hat{A}$ toward $T_{\!+}$ and penalize mass on hair:
\begin{equation}
\mathcal{L}_{\text{att}}
= -\langle T_{\!+},\,\log \hat{A}\rangle
+ \lambda_{\text{neg}}\,
\frac{\langle \hat{A},\,M_{\text{hair}}\rangle}
{\langle \mathbf{1},\,M_{\text{hair}}\rangle+\varepsilon}.
\end{equation}
When parsing masks are absent, this term is omitted.

\subsection{Objective}
\label{ssec:loss}
We optimize a weighted sum of four terms. Here, $\mathcal{L}_{\text{id}}$ and $\mathcal{L}_{\text{tri}}$ are well-known identity classification and triplet losses, $\mathcal{L}_{\text{att}}$ is the parsing-guided attention loss defined in Sec.~\ref{ssec:RPA}, and $\mathcal{L}_{\text{cal}}$ denotes the clothes-adversarial loss adopted from our baseline CAL\cite{gu2022clothes}.


\begin{equation}
\mathcal{L}_{\text{total}}
= \mathcal{L}_{\text{id}}
+ \lambda_{\text{tri}}\,\mathcal{L}_{\text{tri}}
+ \lambda_{\text{att}}\,\mathcal{L}_{\text{att}}
+ \lambda_{\text{cal}}\,\mathcal{L}_{\text{cal}},
\end{equation}
where $\lambda_{\text{tri}},\,\lambda_{\text{att}},\,\lambda_{\text{cal}}$ balance the contributions.

\noindent\textbf{Inference.}
At test time, we only use the ungated $F_{\text{ID}}$ branch, followed by $\ell_2$ normalization and cosine similarity for retrieval.

\vspace{-1ex}
\section{Experiments}
\label{sec:exp}
\subsection{Datasets and Protocols}
\label{ssec:data}
\noindent\textbf{Datasets.}
We evaluate the proposed method on four mainstream Cloth-changing person Re-ID benchmarks: \textbf{PRCC} \cite{yang2019person}, \textbf{LTCC} \cite{qian2020long}, \textbf{VC-Clothes} \cite{wan2020person}, and \textbf{LaST} \cite{shu2021large}. The first three are medium-scale datasets, while \textbf{LaST} is a large-scale dataset.

\noindent\textbf{Implementation details.}
We use \textbf{CAL}\cite{gu2022clothes} as the baseline CC-ReID model in our experiments as it generally performed the best with our generated data across all datasets. We use ResNet-50 pre-trained on ImageNet with a \textbf{maxavg} global pooling head and BNNeck. Inputs are resized to \(384\times192\). For random erasing, the optimal erase range is [0.1, 0.3] on PRCC, LaST and VC-Clothes, and [0.2, 0.5] on LTCC. We train for 60 epochs using Adam (lr \(3.5\times10^{-4}\), weight decay \(5\times10^{-4}\)) with a step scheduler (decay \(\times0.1\) at epochs 20 and 40). All experiments run on two NVIDIA RTX 3080 Ti GPUs. For \textbf{PRCC/LTCC/VC-Clothes} we sample \(4\) identities \(\times\) \(16\) images per identity per GPU, for \textbf{LaST} we sample \(2\) identities \(\times\) \(16\) images per identity per GPU. $\lambda_{\text{cal}}\ $ performs best at 0.5 on PRCC/LaST/LTCC, 1.0 on VC-Clothes, and $\lambda_{\text{att}} = 1.0\ $ performs best across all datasets.  Evaluation is performed every 5 epochs.

\noindent\textbf{Evaluation metrics.}
We report Rank-1 (R1) and mean Average Precision (mAP), the two common metrics in Cloth-changing Re-ID. For all dataset, we evaluate our method under the standard setting and the cloth-changing setting following\cite{wang2024exploring}.

\vspace{-1ex}
\subsection{Comparison with State-of-the-Art Methods}
\label{ssec:sota}
\input{table/table_prcc_ltcc_styled}

\noindent\textbf{Comparative results on PRCC and LTCC.}
Tables~\ref{tab:prcc_ltcc_style} reports comparisons on PRCC and LTCC with classic Re-ID methods (e.g., HACNN~\cite{li2018harmonious}, PCB~\cite{sun2018beyond}, IANet~\cite{hou2019interaction}) and recent CC-ReID methods (e.g., AIM~\cite{yang2023good}, RLQ~\cite{pathak2025coarse}, CAL~\cite{gu2022clothes}).
On \textbf{PRCC (cloth-changing)}, \textbf{MSP-ReID} shows a clear margin over the CAL baseline (about ten points in Rank-1) and is on par with, or slightly better than, recent strong competitors such as RLQ.
On \textbf{PRCC (standard)}, performance saturates at the upper bound.
On \textbf{LTCC (cloth-changing)}, our method yields consistent gains over CAL, while on \textbf{LTCC (standard)} it delivers the best retrieval quality in mAP and competitive Rank-1 among RGB-only approaches.

\noindent\textbf{Comparative results on VC-Clothes.}
Table~\ref{tab:vcclothes_style} compares with conventional baselines (MDLA~\cite{qian2017multi}, PCB~\cite{sun2018beyond}, PS~\cite{shu2021semantic}, FSAM~\cite{hong2021fine}, BSGA~\cite{mu2022learning}, CAL~\cite{gu2022clothes}) and recent CC-ReID lines (STL+ACL~\cite{ding2024improvement}, DLCR~\cite{siddiqui2025dlcr}).under the \textbf{General} and \textbf{SC} protocols, \textbf{MSP-ReID} is competitive with the strongest counterparts.
In the more challenging \textbf{CC} protocol, it clearly improves over CAL and matches the best reported Rank-1, while remaining RGB-only, methods with higher mAP typically rely on tailored designs or auxiliary signals.

\noindent\textbf{Comparative results on LaST.}
Table~\ref{tab:last_style} presents results on the large-scale LaST (CC) benchmark against OSNet~\cite{zhou2019omni}, BoT~\cite{luo2019bag}, mAPLoss~\cite{shu2021large}, MCL~\cite{jin2022meta}, IMS+GEP~\cite{zhao2023joint}, RLQ~\cite{pathak2025coarse}, and CAL~\cite{gu2022clothes}.It indicates that on the large-scale LaST (CC) benchmark, \textbf{MSP-ReID} surpasses classic RGB baselines and its CAL baseline by a noticeable margin, leaving only a small gap to the current top-performing RLQ—likely due to domain shift and inevitable hair/clothes boundary noise at scale.


\vspace{-1ex}
\subsection{Ablation Studies}
\label{ssec:ablation}
\input{table/table_vcclothes_styled.tex}
\input{table/table_last_styled.tex}
\input{table/ablation} 
\begin{figure}[t]
  \centering
  \vspace{-3ex}
  \includegraphics[width=0.88\linewidth,height=0.55\linewidth]{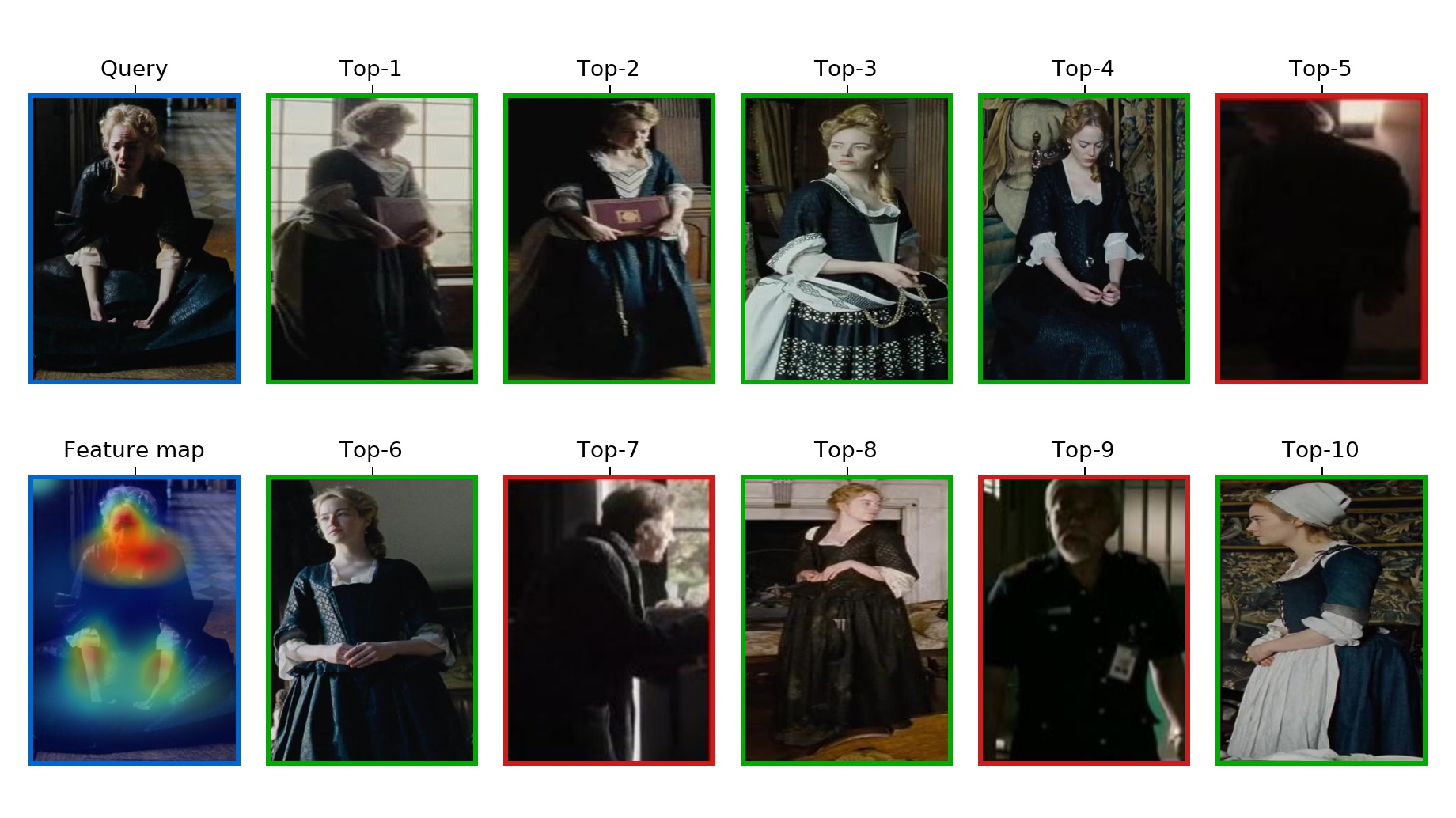}
  \vspace{-4ex}
  \caption{\textbf{Qualitative retrieval and attention.}
  Query (top-left) and its \emph{feature map} (bottom-left) are shown alongside Top-1$\sim$Top-10 retrievals.
  Green boxes are correct matches and red boxes are false.}
  \label{fig:retrieval_top10}
  \vspace{-3.0ex}
\end{figure}

\noindent\textbf{Effectiveness of HSOA.}
As shown in Tab.~\ref{tab:ablation_combo}, removing or adding HSOA clearly changes performance trends. HSOA synthesizes same-ID but different-hairstyle views, explicitly decoupling hairstyle from identity. This breaks the “hair shortcut” and drives the network to focus on identity-related cues (facial structure, exposed-skin geometry, limb proportions) without introducing any extra inputs at test time.

\noindent\textbf{Effectiveness of CPRE.}
Tab.~\ref{tab:ablation_combo} also indicates that CPRE steadily improves robustness. By constructing a raw/erased dual-view with a controllable clothing keep ratio, CPRE suppresses the dominance of large apparel regions while \emph{preserving} partial clothing semantics. This controlled erasing regularizes training and yields more calibrated rankings, rather than overfitting to textures or colors.

\noindent\textbf{Effectiveness of RPA.}
Our Region-based Parsing Attention (RPA) is used only at training time to reweight regions—boosting face/limbs and suppressing hair. As reflected by the ablations in Tab.~\ref{tab:ablation_combo}, aligning attention with identity-related anatomy reduces gradients attracted to hair/clothes boundaries and complements the above augmentations, all while keeping inference RGB-only.

\noindent\textbf{Module collaboration.}
The table further shows that single modules are helpful but suboptimal. Pairwise combinations consistently perform better, revealing clear complementarity: HSOA changes the \emph{data} (hairstyle diversity), CPRE changes the \emph{appearance reliability} (keep-ratio erasing), and RPA changes the \emph{feature selection} (region weighting). Using all three together fully exploits these effects and yields the most stable cross-clothes/hairstyle retrieval.

\noindent\textbf{Visualization of retrieval results.}
Fig.~\ref{fig:retrieval_top10} qualitatively verifies these behaviors. Compared with the baseline, our method correctly retrieves the same identity under clothing and hairstyle changes, avoiding confusion from similar colors or textures. The feature maps exhibit a clear “hot face, cold hair/clothes” pattern—high responses on face, shoulders, and limbs with suppressed activations on hair and large apparel regions—consistent with HSOA, CPRE, and RPA.

\vspace{-1ex}
\section{Conclusion}

In this paper, we proposed MSP-ReID, a unified framework that mitigates hairstyle-induced bias and preserves structural information for cloth-changing person re-identification. MSP-ReID introduces Hairstyle-Oriented Augmentation (HSOA) to generate same-identity images with diverse hairstyles, explicitly reducing reliance on hairstyle cues. Cloth-Preserved Random Erasing (CPRE) selectively removes pixels within clothing regions while retaining geometric and contextual information, thereby alleviating texture bias. Region-based Parsing Attention (RPA) further leverages parsing priors to enhance identity-relevant regions and suppress hair features during training. Extensive experiments on PRCC, LTCC, VC-Clothes, and LaST confirm that MSP-ReID achieves state-of-the-art performance and exhibits strong robustness to both clothing and hairstyle changes, highlighting its potential for practical, long-term CC-ReID applications.


\bibliographystyle{IEEEtran}
\bibliography{strings,main}

\end{document}

%% file: table/table_prcc_ltcc_styled.tex

\begin{table}[t]
\centering
\setlength{\tabcolsep}{4pt}
\vspace{-2ex}
\caption{Results on PRCC and LTCC. The best result is in bold, and the second-best result is underlined.}
\vspace{2ex}
\resizebox{0.9\columnwidth}{!}{%
\begin{tabular}{l c cc cc cc cc} 
\toprule
\multirow{3}{*}{\textbf{Methods}} & \multirow{3}{*}{\textbf{Year}} &
\multicolumn{4}{c}{\textbf{PRCC}} &
\multicolumn{4}{c}{\textbf{LTCC}} \\
& & \multicolumn{2}{c}{Cloth-Changing} & \multicolumn{2}{c}{Standard} &
      \multicolumn{2}{c}{Cloth-Changing} & \multicolumn{2}{c}{Standard} \\
\cmidrule(lr){3-4} \cmidrule(lr){5-6} \cmidrule(lr){7-8} \cmidrule(lr){9-10}
& & R1 & mAP & R1 & mAP & R1 & mAP & R1 & mAP \\
\midrule

HACNN \cite{li2018harmonious} & CVPR '18 & 21.8 & - & 82.5 & - & 21.6 & 9.3 & 60.2 & 26.7 \\
PCB \cite{sun2018beyond} & ECCV '18 & 41.8 & 38.7 & 99.8 & 97.0 & 23.5 & 10.0 & 65.1 & 30.6 \\
IANet \cite{hou2019interaction} & CVPR '19 & 46.3 & 46.9 & 99.4 & 98.3 & 25.0 & 12.6 & 63.7 & 31.0 \\
FSAM \cite{hong2021fine} & CVPR '21 & 54.5 & - & 98.8 & - & 38.5 & 16.2 & 73.2 & 35.4 \\
AIM \cite{yang2023good} & CVPR '23 & 57.9 & 58.3 & 100.0 & \underline{99.9} & 40.6 & 19.4 & 76.3 & 41.1 \\
CCFA \cite{han2023clothing} & CVPR '23 & 61.2 & 58.4 & 99.6 & 98.7 & 45.3 & 22.1 & 75.8 & 42.5 \\
Instruct-ReID \cite{he2024instruct} &  CVPR'24 & 54.2 & 52.3 & - & - & - & - & 75.8 & 52.0 \\
LIFTCAP \cite{xiong2024cloth} & TVT'24 & 54.3 & 55.6 & 100.0 & 99.8 & 37.0 & \textbf{39.7} & - & - \\
JIMGP \cite{zhao2023joint} & TMM'24 & 57.3 & \textbf{65.8} & 99.7 & 99.8 & 43.4 & 18.2 & - & - \\
CISupNet \cite{hu2025causal} & ICASSP'25 & 58.3 & 58.2 & 100.0 & 99.8 & 41.5 & 19.2 & 76.0 & 41.6 \\
FAIM \cite{zhao2025clothes} & TMM'25 & 59.8 & 62.5 & 100.0 & \textbf{100.0} & \textbf{48.2} & \underline{27.5} & \textbf{79.5} & 53.4 \\
RLQ \cite{pathak2025coarse} & arXiv'25 & 64.0 & 63.2 & 100.0 & 99.8 & \underline{46.4} & 21.5 & 76.9 & 41.8 \\
\midrule
CAL \cite{gu2022clothes} & CVPR '22 & 55.2 & \underline{55.8} & 100.0 & 99.8 & 40.1 & 18.0 & 74.2 & 40.8 \\
\rowcolor{gray!10} \textbf{ours} & - & \textbf{65.1} & \underline{63.4} & \textbf{100.0} & 99.1 & 41.6 & 19.3 & \underline{78.7} & \textbf{60.1} \\
\bottomrule
\end{tabular}%
} 
\vspace{-2ex}
\label{tab:prcc_ltcc_style}
\end{table}

%% file: table/table_vcclothes_styled.tex
\begin{table}[t]
\vspace{-2ex}
\caption{Results on VC-Clothes under General, SC and CC protocols. The best and second-best results are shown in bold and underlined.}
\vspace{1.5ex}
\centering

\resizebox{0.9\columnwidth}{!}{%
\begin{tabular}{@{}l c cc cc cc@{}}
\toprule
\multirow{2}{*}{Method} & \multirow{2}{*}{Year} &
\multicolumn{2}{c}{General} & \multicolumn{2}{c}{SC} & \multicolumn{2}{c}{CC} \\
& & R1 & mAP & R1 & mAP & R1 & mAP \\
\midrule
MDLA \cite{qian2017multi} & ICCV'17 & 88.9 & 76.8 & 94.3 & 93.9 & 59.2 & 60.8 \\
PCB \cite{sun2018beyond} & ECCV'18 & 87.7 & 74.6 & 94.7 & 94.3 & 62.0 & 62.2 \\
PS \cite{shu2021semantic} & SPL'21 & 93.1 & 84.9 & 94.7 & 92.9 & 82.4 & 80.3 \\
FSAM \cite{hong2021fine} & CVPR'21 & - & - & 94.7 & 94.8 & 78.6 & 78.9 \\
BSGA \cite{mu2022learning} & BMVC'22 & \underline{94.4} & \textbf{88.2} & 94.9 & 94.4 & \underline{84.5} & \underline{84.3} \\
STL and ACL \cite{ding2024improvement} & CSCWD'24 & - & - & 95.1 & \underline{95.5} & 83.9 & \textbf{85.2} \\
PAH-Net\cite{thanh2024enhancing} & IJON'24 & 94.2 & 90.2 & \underline{95.3} & 95.2 & 86.4 & 86.1 \\
DLCR \cite{siddiqui2025dlcr} & WACV'25 & - & - & - & - & \textbf{87.1} & 81.1 \\
\midrule
CAL \cite{gu2022clothes} & CVPR'22 & 92.9 & 87.2 & 95.1 & 95.3 & 81.4 & 81.7 \\
\rowcolor{gray!10}\textbf{Ours} & — & \textbf{94.5} & \underline{87.2} & \textbf{95.4} & \textbf{95.6} & \textbf{87.1} & 82.3 \\
\bottomrule
\end{tabular}}
\vspace{-3ex}
\label{tab:vcclothes_style}
\end{table}

%% file: table/table_last_styled.tex
\begin{table}[t]
\centering
\setlength{\tabcolsep}{4pt}

\caption{Results on LaST (CC protocol). The best result is in bold, and the second-best result is underlined.}
\vspace{2ex}

\resizebox{0.5\columnwidth}{!}{%
\begin{tabular}{l c c c} 
\toprule
\textbf{Methods} & \textbf{Year} & \textbf{R1} & \textbf{mAP} \\
\midrule
OSNet \cite{zhou2019omni}          & ICCV'19   & 63.8 & 20.9 \\
BoT  \cite{luo2019bag}          & CVPR'19   & 68.3 & 25.3 \\
mAPLoss \cite{shu2021large}        & arXiv'21  & 69.9 & 27.6 \\
MCL \cite{jin2022meta}           & MM’22     & 75.0 & 22.7 \\
Lu and Jin \cite{lu2023dual}    & AIMS’23   & 68.9 & 24.1 \\
IMS+GEP \cite{zhao2023joint}        & TMM’23    & 73.2 & 29.8 \\
RLQ \cite{pathak2025coarse}            & arXiv'25  & \textbf{77.9} & \textbf{35.3} \\
\midrule
CAL \cite{gu2022clothes} & CVPR'22  & 73.7 & 28.8 \\
\rowcolor{gray!10}\textbf{ours} & - & \underline{75.4} & \underline{30.6} \\
\bottomrule
\end{tabular}%
} 
\vspace{-3ex}
\label{tab:last_style}
\end{table}

%% file: table/ablation.tex
\begin{table}[t]
\centering

\setlength{\tabcolsep}{4pt}
\vspace{-2ex}
\caption{Ablation study of HSOA, CPRE, PRA on PRCC and VC-Clothes.}
\vspace{1ex}

\resizebox{0.86\columnwidth}{!}{%
\begin{tabular}{c c c c c cc cc} 

\toprule
\textbf{Methods} & \textbf{Baseline} & \textbf{HSOA} & \textbf{CPRE} & \textbf{PRA} &
\multicolumn{2}{c}{\textbf{PRCC}} &
\multicolumn{2}{c}{\textbf{VC-Clothes}} \\
\cmidrule(lr){6-7}\cmidrule(lr){8-9}
& & & & &
\multicolumn{1}{c}{\textbf{R1}} & \multicolumn{1}{c}{\textbf{mAP}} &
\multicolumn{1}{c}{\textbf{R1}} & \multicolumn{1}{c}{\textbf{mAP}} \\
\midrule
1 & \checkmark &            &            &            & 55.2 & 55.8 & 81.4 & 81.7 \\
2 & \checkmark & \checkmark &            &            & 61.0 & 59.6 & 84.9 & 82.0 \\
3 & \checkmark &            & \checkmark &            & 56.7 & 57.0 & 82.4 & 81.9 \\
4 & \checkmark &            &            & \checkmark & 57.7 & 57.8 & 82.2 & 81.8 \\
5 & \checkmark & \checkmark & \checkmark &            & 63.1 & 61.1 & 86.4 & 82.2 \\
6 & \checkmark & \checkmark &            & \checkmark & 64.3 & 62.1 & 86.2 & 82.1 \\
7 & \checkmark &            & \checkmark & \checkmark & 59.5 & 59.3 & 83.4 & 82.1 \\
\midrule
\rowcolor{gray!10}\textbf{Ours} &
\checkmark & \checkmark & \checkmark & \checkmark &
\textbf{65.1} & \textbf{63.4} & \textbf{87.1} & \textbf{82.3} \\
\bottomrule
\end{tabular}%
}

\label{tab:ablation_combo}
\end{table}